\def\year{2022}\relax
\documentclass[letterpaper]{article} 
\usepackage{aaai21}  
\usepackage{times}  
\usepackage{helvet} 
\usepackage{courier}  
\usepackage[hyphens]{url}  
\usepackage{graphicx} 
\usepackage{amsmath}
\usepackage{amsfonts}
\usepackage{latexsym}
\usepackage{multirow}
\usepackage{tabularx}
\usepackage{booktabs}
\usepackage{url}
\usepackage{hyperref}
\usepackage{color}
\usepackage{CJKutf8}

\newcommand{\Zh}[1]{
\begin{CJK}{UTF8}{gbsn}#1\end{CJK}}

\usepackage{natbib}  
\usepackage{caption} 
\title{Query-Response Interactions by Multi-tasks in Semantic Search for Chatbot Candidate Retrieval}
\author {
    Libin Shi,
    Kai Zhang,
    Wenge Rong,
}
\affiliations {

    olivier.slb@outlook.com, zhangkai.plus@foxmail.com, w.rong@buaa.edu.cn
}
\begin{document}

\maketitle

\begin{abstract}
Semantic search for candidate retrieval is an important yet neglected problem in retrieval-based Chatbots, which aims to select a bunch of candidate responses efficiently from a large pool. The existing bottleneck is to ensure the model architecture having two points: 1) rich interactions between a query and a response to produce query-relevant responses; 2) ability of separately projecting the query and the response into latent spaces to apply efficiently in semantic search during online inference. To tackle this problem, we propose a novel approach, called Multitask-based Semantic Search Neural Network (MSSNN) for candidate retrieval, which accomplishes query-response interactions through multi-tasks. The method employs a Seq2Seq modeling task to learn a good query encoder, and then performs a word prediction task to build response embeddings, finally conducts a simple matching model to form the dot-product scorer. Experimental studies have demonstrated the potential of the proposed approach.
\end{abstract}

\section{Introduction}

Recent years have witnessed a great development of non-task oriented chatbots. Existing approaches fall into generation-based methods~\cite{DBLP:conf/acl/ShangLL15,DBLP:conf/aaai/SerbanSBCP16,DBLP:conf/aaai/XingWWHZ18}, and retrieval-based methods~\cite{DBLP:conf/nips/HuLLC14,DBLP:conf/sigir/YanSW16,DBLP:conf/acl/WuWXZL17}. The retrieval-based approaches, which enjoys informative and fluent responses, have attracted more attention in industry, e.g., social chatbot XiaoIce~\cite{DBLP:journals/corr/abs-1812-08989} and E-commerce assistant AliMe~\cite{DBLP:conf/cikm/LiQCWGHRZZWJC17}. 

The pipeline of retrieval-based chatbots includes two modules: a retrieval module and a re-rank module. The retrieval module searches a number of candidates from a large permitted corpus efficiently, then these candidates are fed to the re-rank module for further selection. 
A common practice of the retrieval module is to build a keyword-based index for conversational query-response (QR) pairs. Given a user input, the index is able to find several candidates in a short time based on the keyword overlap between the input and indexed queries. This approach, however, suffers from two limitations: 1) for many specific domains, a large amount of high-quality responses are organized in unstructured documents, so it is intractable to collect such QR pairs, as such it is not suitable to consider matching features calculated by online query and local collected queries, like \cite{DBLP:journals/corr/JiLL14}. 2) keyword-based retrieval method performs poorly when computing the similarity between indexed responses and a user input, since a good response sometimes does not share the same words with the given user input, which is analyzed in Section Results and analysis. 

\begin{figure}[!t]
  \centering
  \includegraphics[width=0.6\linewidth]{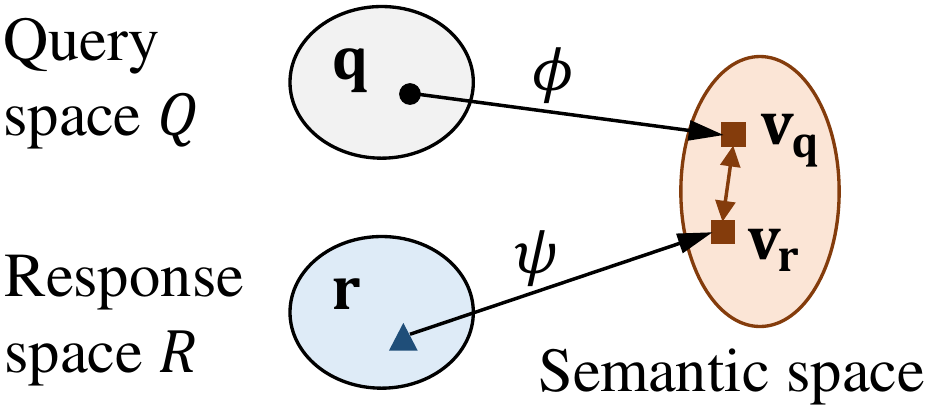}
  \caption{Query-response semantic search definition}
  \label{fig:matching}
\end{figure}

Hence, we employ the semantic search technique for response retrieval. Specifically, a model that is capable of separately projecting queries and responses into the distributed representations, as shown in Fig. \ref{fig:matching}, should be trained at the first step. When a query comes at the online inference step, an index that stores all responses and their vectors retrieves similar responses efficiently according to the Euclidean distances between the corresponding vectors of the query and the responses.  

Consequently, how to learn the project function plays a vital role in semantic search, which should find a trade-off between accuracy and latency. 
Previous QR matching models are roughly categorized into two types: matching function learning and representation learning. Matching function learning \cite{DBLP:conf/aaai/PangLGXWC16,DBLP:conf/ijcai/WanLXGPC16,DBLP:conf/www/PengCXLZL20} carries out QR interaction at the beginning of the architecture which is unable for semantic searching as these methods have query and response interactions resulting in a huge latency in retrieval. Representation learning~\cite{DBLP:conf/acl/WangN15,DBLP:journals/corr/TanXZ15,DBLP:journals/taslp/PalangiDSGHCSW16} does not have this problem. However, these methods learn the representations of the query and the response without rich interactions of each other; What's worse, semantic matching generally uses negative sampling to construct training data hence yielding large training-inference bias because it should select a relevant response from a large number of corpus during online application, while it merely contains a certain number of negative samples during training. As such these two problems lead to shallow representations of the query and the response. In the worst-case, the retrieval pipeline even returns totally irrelevant responses with these models. It is worth noting that Google's e-mail reply with response suggestions is an attempt of using improved matching model \cite{DBLP:journals/corr/HendersonASSLGK17}, and yet it merely has a small response candidates to select and our experiments also show that it could not be a good practice for daily conversational chatbots. Some recent works based on Bert try to apply semantic search in some tasks, like similar text matching \cite{DBLP:conf/emnlp/ReimersG19} or open-domain question answering \cite{DBLP:conf/acl/LeeCT19}, but they also suffer the same problems with above works when being applied in this task.

To overcome the issues that call for both ensuring QR interaction and forming dot-product scoring at the end, we involve the multi-task learning framework~\cite{DBLP:conf/icml/CollobertW08}, which enforces QR interactions in extra tasks in addition to the main dot-product scoring task. Thus we propose a novel model called Multitask-based Semantic Search Neural Network (MSSNN) for candidates retrieval. Our model contains three parts: 1) first is \textbf{Seq2Seq} which is utilized to learn a good encoder for query's representation; 2) second is \textbf{Word Predictor}, it is designed for mapping query's representation to response space as well as realizing response embeddings; 3) third is a simple \textbf{Matching Model} which is employed to form the final dot-product scoring. 

Our contributions include: 1) we explore the possibility of using semantic search instead of Lucene index for candidate retrieval in chatbots; 2) we elaborate a multitask-based model to utilize the extra tasks to emphasize interactions between query-response pairs as well as to accomplish the semantic representations of them separately; 3) the experimental studies have verified the effectiveness of our proposed model. The implementation code will be published.

\section{Problem Definition}\label{sec:problem}

Suppose we have a dataset $\mathcal{D} = \{\mathbf{q}_i,\mathbf{r}_i\}_i^N$, where $\mathbf{q} = \{ q_1 \ldots q_n \}$ is a conversational query and $\mathbf{r}_i = \{r_1 \ldots r_m\}$ is its corresponding response. Our goal is to learn two mapping functions $\phi(\cdot)$ and $\psi(\cdot)$, that are capable of projecting $\mathbf{q}$ and $\mathbf{r}$ into the same latent space $\mathbb{R}^d$. Given a query $\mathbf{q}_i$, we hope the relevant response $\mathbf{r}_i$ is closer to it than other irrelevant responses in the latent space, which is formulated as  $ \phi(\mathbf{q}_i) \cdot \psi(\mathbf{r}_i) > \phi(\mathbf{q}_i) \cdot \psi(\mathbf{r}_j), \forall i, j,  j \neq i $.

By this means, in the online phase, we can project all responses\footnote{Some responses are extracted from plain text, which do not have an antecedent query.} $R=\{\mathbf{r}_i\}_i^N$ into the latent space in advance $\psi(R)$. 
Afterwards, given a user query $\mathbf{q}$, the system returns top-k relevant responses by calculating the dot-product between the response representations and $\phi(\mathbf{q})$. In practice, the top-k nearest neighbors can be found by some approximate algorithms, such as kd-tree~\cite{DBLP:conf/iccv/ShenLZYS15} and local sensitive hashing \cite{DBLP:conf/uai/Shrivastava015}, which could accelerate the retrieval speed significantly. 

The problem of response retrieval is different from response selection, since the target of response retrieval is the speed. It requires to learn query and response representation separately, illustrated in Fig. \ref{fig:matching}, thus, popular matching function learning models \cite{DBLP:conf/acl/WuWXZL17} are not suitable for this task. In following sections, we will elaborate how to learn mapping functions $\phi$ and $\psi$.

\section{Approach}
\begin{figure}[!t]
  \centering
  \includegraphics[width=1.02\linewidth]{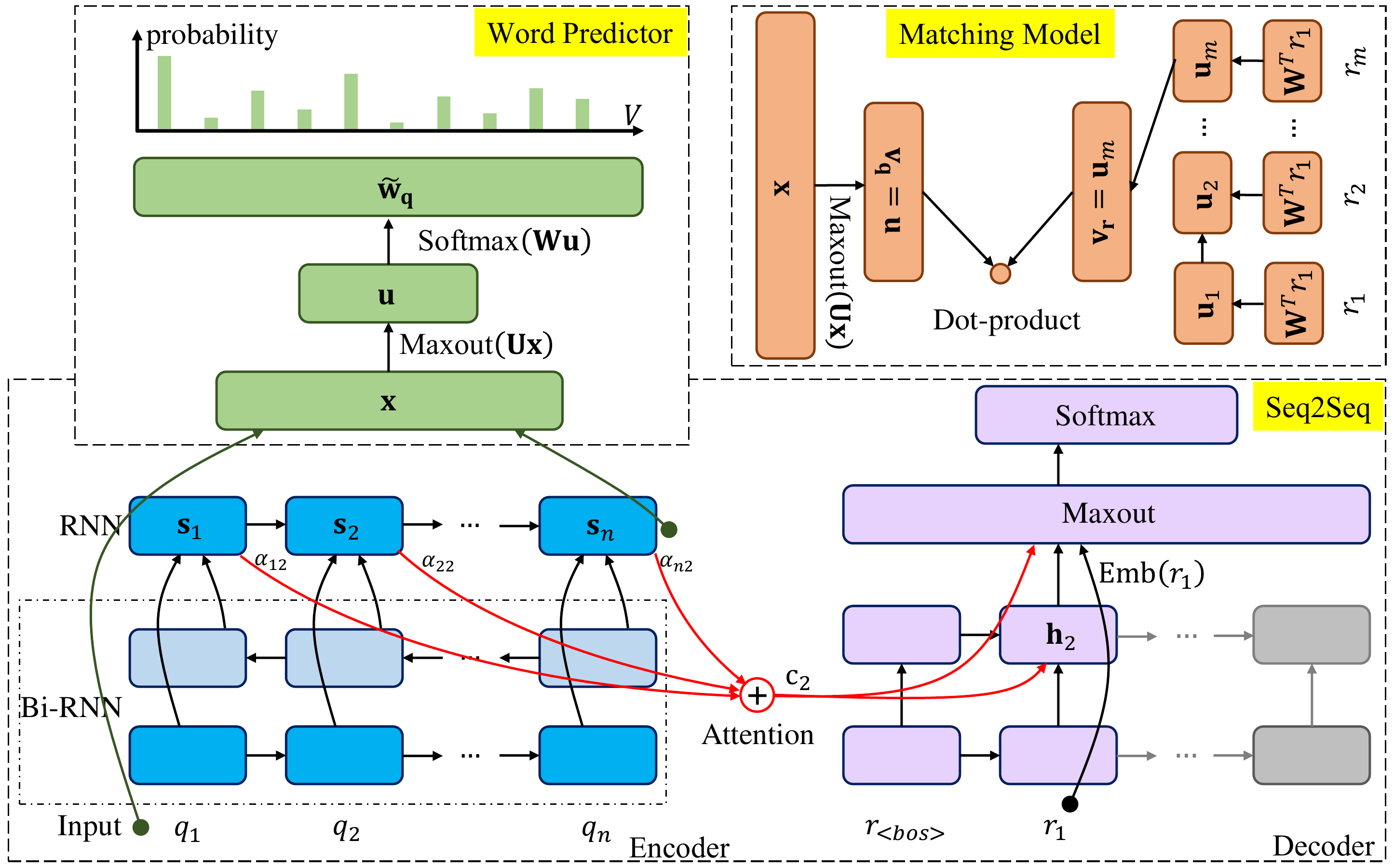}
  \caption{Model architecture of Multitask-based Semantic Search Neural Network.}
  \label{fig:mt4ss}
\end{figure}

To formalize two mapping functions $\phi$ and $\psi$ at the same time to ensure QR interactions, this section presents a novel model, called Multitask-based Semantic Search Neural Network (MSSNN). The architecture is shown in Fig. \ref{fig:mt4ss}, which includes three parts, \textbf{Seq2Seq}, \textbf{Word Predictor} and \textbf{Matching Model}. 

\subsection{Seq2Seq for Query Encoder}\label{sec:seq2seq}

Motivated by the modeling ability of Seq2Seq on the the human-human conversations, we setup a Seq2Seq task in our MSSNN model instead of only using its encoder. One of our experimental studies, in Section First-step prediction in Seq2Seq, has shown that after training a single generation task with Seq2Seq model on QR-pair datasets, if we pass a query into the encoder and let decoder predict top-5000 words from the vocabulary at the first decoding step, these predictive top-5000 words could coverage 93\% words of the corresponding response. The first-step prediction merely depends on the input query but could predict mostly words in the response, thus it shows the query encoder of Seq2Seq has learned a good connection between queries and responses. 



Driven by the above observation, we setup the Seq2Seq task on conversational $(\mathbf{q}, \mathbf{r})$ pairs for training a query encoder, as shown in Fig. \ref{fig:mt4ss}.
The encoder has two layers, the first layer is a bidirectional RNNs (BiRNNs) and then is a single RNN. Given $\mathbf{q}$, outputs of the BiRNNs layer is,
\begin{equation}
[\overrightarrow{\mathbf{s}^0}, \overleftarrow{\mathbf{s}^0}] = \mathrm{BiRNNs}(\mathbf{q})
\end{equation}
where $\overrightarrow{\mathbf{s}^0}$ and $\overleftarrow{\mathbf{s}^0}$ denote respectively outputs of forward and reverse RNNs. Then we concatenates them as input of the top layer RNN, $\mathbf{s}^0 = (\overrightarrow{\mathbf{s}_1^0} \oplus \overleftarrow{\mathbf{s}_1^0}, \dots, \overrightarrow{\mathbf{s}_n^0} \oplus \overleftarrow{\mathbf{s}_n^0})$, and the outputs are obtained by hidden states of this RNN,
\begin{equation}
\mathbf{s}_{|\mathbf{s}_1, \dots, \mathbf{s}_n}  = \mathrm{RNN}(\mathbf{s}^0)
\end{equation}

The decoder has two RNN layers with Bahdanau attention, and its decoding architecture and formula follow the work by~\cite{DBLP:journals/corr/BahdanauCB14}. 

The conditional probability at each decoding step in the joint probability $P(\mathbf{r}|\mathbf{q}) = \prod_1^m{P(r_i|r_{i-1}, \mathrm{\textit{context}})}$ is calculated as follows,
\begin{equation}
P(r_i|r_{i-1}, \mathrm{\textit{context}}) = \mathrm{Softmax}(t_i)
\end{equation}
where $\mathrm{\textit{context}}$ stands for the $\mathbf{q}$ as well as the historical tokens in $\mathbf{r}$, and $t_i$ is the output logits from the $\mathrm{Maxout}$ as described in \cite{DBLP:journals/corr/BahdanauCB14}. Finally, the Seq2Seq model is trained to maximize this conditional probability or to minimize the negative log-likelihood (NLL),
\begin{equation}\label{eq:loss_nll}
    L_{\mathrm{NLL}} = - \sum_1^m{\log P(r_i|r_{i-1}, \mathrm{\textit{context}})}
\end{equation}

We can see from the loss function in Eq. \ref{eq:loss_nll} that each token $r_i$'s prediction of the response should depend on the query, which conducts NLL loss in the Seq2Seq task is a strong interaction between the query and the response. In addition, the predicting process attempts to search the right token from the entire vocabulary at each decoding step, while the matching model only seeks the right response from a certain number of negative samples. As a result, Seq2Seq model is executed as an extra task for our semantic search.


\subsection{Word Predictor for Response Embedding} \label{sec:word_predictor}
In order to map query representations to response latent space as well as to acquire response embedding, Word Predictor is designed to use the outputs of query encoder, which is described in Section Seq2Seq for Query Encoder, to predict all the words in the corresponding response. This idea is inspired by using Skip-gram model to train word2vec, which uses the center word to predict all the context \cite{DBLP:journals/corr/MikolovLS13} and yields word embeddings. Therefore, we use query words as center words to predict all the words in the corresponding response as such to yield response embedding that can map response words to query representation space.


For this, a new manner of data construction is introduced. For each distinct query, $\mathbf{q}$, we collect all words in its related responses and represent them by bag-of-words, $\mathbf{w_q} \in \mathbb{R}^{|\mathcal{V}|}$, where $|\mathcal{V}|$ is vocabulary size of responses. $\mathbf{w_q}$ is a discrete representation of all related words for each query, but we need to transform it into probability distribution. Thus we utilize the normalized TF-IDF, $\pi(\mathbf{w_q}) \in \mathbb{R}^{|\mathcal{V}|}$, as targets of Word Predictor. The training data is finally organized into $(\mathbf{q}, \pi(\mathbf{w_q}))$, and the architecture is illustrated in Fig. \ref{fig:mt4ss}.

It first utilizes the query encoder from the Seq2Seq framework, and concatenates the average embeddings and the average encoder's outputs,

\begin{equation}\label{eq:wordpred_x}
  \mathbf{x} = \frac{1}{n} \sum_{i <= n}{\mathrm{Emb}(q_i)} \oplus \frac{1}{n} \sum_{i <= n}{\mathrm{s}_i}
\end{equation}
where $\mathbf{x} \in \mathbb{R}^l$. The following is a dense layer with $\mathrm{Maxout}$ activation function,
\begin{equation}\label{eq:wordprd_v}
  \mathbf{u} = \mathrm{Maxout}(\mathbf{U} \mathbf{x})
\end{equation}
where $\mathbf{U} \in \mathbb{R}^{d \times l}$. Word-predicting distributed probability, $\mathbf{\tilde{w}_q}$, is calculated as follow:
\begin{equation}\label{eq:wordpred_w}
  \mathbf{\tilde{w}_q} = \mathrm{Softmax}(\mathbf{W} \mathbf{u})
\end{equation}
where $\mathbf{W} \in \mathbb{R}^{|\mathcal{V}| \times d}$. And finally, loss function $L_{\mathrm{KL}}$ is set to minimize the KL distance between the prediction $\mathbf{\tilde{w}_q}$ and the targets $\pi(\mathbf{w_q}) \in \mathbb{R}^{|\mathcal{V}|}$ .

\begin{equation}\label{eq:ce_wp}
  L_{\mathrm{KL}} = \mathrm{KL}(\mathbf{\tilde{w}_q} || \pi(\mathbf{w_q})) 
\end{equation}

Above the encoder of Seq2Seq, Word Predictor focuses on optimizing $\mathbf{U}$ and $\mathbf{W}$. Notice that the function of $\mathbf{W}$ is to project from query representation, $\mathbf{u} \in \mathbb{R}^d$, to response words vocabulary by minimizing KL distance. Thereby its transposition, $\mathbf{W}^T$, can be used as response embeddings to generate word-level representations of responses.

\subsection{Matching Model for Scorer}
Matching Model operates finally with the help of the above two parts, and its architecture is designed for forming the final dot-product scorer, which is shown in Fig. \ref{fig:mt4ss}. It also accomplishes the formula of the two mapping functions $\phi$ and $\psi$. The training data is constructed as $(\mathbf{q}, \mathbf{r}^+, \mathbf{r}^-)$, where $\mathbf{r}^-)$ stands for randomly sampled negative samples.

It first constitutes the query-representative module, using the Seq2Seq's encoder and the part of Word Predictor. Afterwards by inputting the query $\mathbf{q}$ and feeding it forward to the semantic representation through the formulas Eq. \ref{eq:wordpred_x} and Eq. \ref{eq:wordprd_v}, and we can obtain its representation, $\mathbf{v_q} = \mathbf{u}$.

Afterwards, to build the response-representative module, it mainly utilizes the response embedding $\mathbf{W}^T \in \mathbb{R}^{d \times |\mathcal{V}|}$ in Eq. \ref{eq:wordpred_w}. Each token, $r_i \in \mathbf{r}$, looks up $\mathbf{W}^T$ and gets its embedding, $\mathbf{W}^T r_i$. Then, it employs a simple RNN (hidden size $=d$) to input these word embeddings, $\mathbf{W}^T r_1 \ldots \mathbf{W}^T r_m$, and to output the last hidden state of RNN as the final representation of this response,
\begin{equation}\label{eq:match_phi}
  \begin{split}
    \mathbf{u}_1, \dots, \mathbf{u}_m &= \mathrm{RNN}(\mathbf{W}^T r_1, \dots \mathbf{W}^T r_m) \\
    \mathbf{v_r} &= \mathbf{u}_m
  \end{split}
\end{equation}
Then we use ranking loss as loss function to further shorten the semantic distance between query representations and response representations,
\begin{equation}\label{eq:ranking_loss}
\resizebox{.866\hsize}{!}{$L_{\mathrm{R}} = \max(0, 1 - \cos(\mathbf{v_q}, \mathbf{v}_{\mathbf{r}^+}) + \cos(\mathbf{v_q}, \mathbf{v}_{\mathbf{r}^-}))$}
\end{equation}
To build the two mapping for forming the dot-product scoring, we define, $\phi(\mathbf{q}) = \frac{\mathbf{v_q}}{||\mathbf{v_q}||}$ and $\psi(\mathbf{r}) = \frac{\mathbf{v_r}}{||\mathbf{v_r}||}$, thus, $\phi(\mathbf{q}) \cdot \psi(\mathbf{r})$ equals to the cosine similarity in the training phase. Finally, the two mapping functions $\phi$ and $\psi$ are formalized by Matching Model.


\subsection{Model training}
To learn the three tasks in MSSNN, the training data is reorganized into $(\mathbf{q}, \mathbf{r}^+, \mathbf{r}^-, \pi(\mathbf{w_q}))$ from the original $(\mathbf{q}, \mathbf{r})$ pairs, where $\mathbf{r}^+$ is ground-truth response to the query while $\mathbf{r}^-$ is a random sampled negative response, and $\pi(\mathbf{w_q})$ is normalized TF-IDF value. And total loss function for the MSSNN is the union of the three loss functions, Eq. \ref{eq:loss_nll}, \ref{eq:ce_wp}, \ref{eq:ranking_loss},
\begin{equation}
L = \alpha L_{\mathrm{NLL}} + \beta L_{\mathrm{\mathrm{KL}}} + \gamma L_{\mathrm{R}}
\end{equation}
where $\alpha, \beta, \gamma$ are hyper-parameters and are all configured as 1 in our experiments. Meanwhile, we set representative feature dimension $d=512$, and set respectively embedding and RNN hidden size as $|E| = 512$ and $|H| = 1024$, totally yielding 74 million parameters. In addition, in our implementation we use gated recurrent units~\cite{DBLP:conf/ssst/ChoMBB14} as RNN model and use Adam optimizer \cite{DBLP:journals/corr/KingmaB14} with learning rate $\mathrm{lr}=0.0002$. And our implementation code will be published.

\subsection{Model Inference}
The entire architecture of MSSNN in Fig. \ref{fig:mt4ss} is used in training phase. At inference time, we first use $\psi$ to project all the permitted corpus into vectors offline, and during online application we use $\phi$ to project a query into a vector and calculate their similarity distance by dot-producting. This vector search procedure can be speed-up by algorithms of approximate nearest neighbor \cite{DBLP:conf/iccv/ShenLZYS15} using open-source tools ~\footnote{\url{https://github.com/spotify/annoy}}.

\section{Experiments}
In this section, we conduct experimental studies to analyze the performance of our proposed model.
\subsection{Setup}
\subsubsection{Dataset} consists of query-response pairs in open-domain conversation from two sources. One is collected from Douban group\footnote{\url{https://www.douban.com/group}}. After cleaning it lefts 10 million query-response pairs, and we randomly select 0.1 million pairs as testing data. Another is Ubuntu Corpus~\cite{DBLP:conf/sigdial/LowePSP15}, which is processed by the script\footnote{\url{https://github.com/rkadlec/ubuntu-ranking-dataset-creator}}, whose vocabulary size is set as $30,000$. Notice that Douban testing data and their vector representations will be published.

\subsubsection{Baselines} include 1) Lucene: build Lucene index on responses and use queries to retrieve. 2) Embedding: average word embeddings pre-trained by Word2Vec\footnote{\url{code.google.com/p/word2vec}}
~\cite{DBLP:conf/nips/MikolovSCCD13}. 3) RNN: two RNNs encode respectively queries and responses to obtain their representative vectors. 4) SA-BiRNNs: bidirectional RNN with self attention~\cite{DBLP:conf/iwcs/RuckleG17}. 5) QA-BiRNNs: query attentive bidirectional RNN~\cite{DBLP:journals/corr/TanXZ15}. In addition to SA-BiRNNs, it adds to utilize outputs of the query encoder as memory to annotate response's encoding. 6) GRSM: Google's response suggestion model for Smart Reply system~\cite{DBLP:journals/corr/HendersonASSLGK17}. 7) Bert \cite{DBLP:journals/corr/abs-1810-04805} has demonstrated to be a powerful language representations, and we fine-tune Bert in this semantic search task by using the architecture introduced in \cite{DBLP:conf/emnlp/ReimersG19} and \cite{DBLP:conf/acl/LeeCT19}, where BERT-Base-Chinese 
 is for Douban dataset and BERT-Base-Uncased 
 is for Ubuntu dataset. To compare fairly, we set dimension of representation vector $\mathbf{v_q}$ and $\mathbf{v_r}$ as $d=512$ of all the above methods except for Bert which is 768.

\begin{table*}[!t]
\centering
\caption{Models comparison on Douban and Ubuntu dataset using metrics of R-precision and R@k.}
\label{tab:comarison}
\begin{tabular}{|l|c|c|c|c|c|c|}
\toprule
Datasets     &\multicolumn{3}{c|}{Douban} & \multicolumn{3}{c|}{Ubuntu} \\
Methods     & R-precision & R@100      & R@1000      & R-precision & R@10      & R@100  \\
\midrule   

Lucene     &  0.0006  & 0.0109  &  0.0925 &  0.0003 & 0.0008  & 0.0056 \\
Embedding   & 0.0001      & 0.0010     & 0.0099      & 0.0009      & 0.0008    & 0.0067       \\
RNN         & 0.0001   & 0.0012    & 0.0126          & 0.0005   & 0.0005    & 0.0052  \\
SA-BiRNNs   & 0.0002   & 0.0017    & 0.0150          & 0.0007   & 0.0006    & 0.0070  \\
QA-BiRNNs   & 0.0008   & 0.0103    & 0.0721          & 0.0030   & 0.0011    & 0.0094       \\

GRSM        & 0.0011   & 0.0114    & 0.0767          & 0.0073   & 0.0014    & 0.0107  \\
Bert        & \textbf{0.0199}   & 0.0809   & 0.2020  & 0.0121   & 0.0203    & 0.0994  \\
MSSNN       & 0.0176   & \textbf{0.1030}    & \textbf{0.2676}     & \textbf{0.0160}   & \textbf{0.0343}    & \textbf{0.1442}  \\
\bottomrule
\end{tabular}
\end{table*}

\begin{table*}[!t]
\centering
\caption{Models comparison on Douban dataset w.r.t embedding-based metrics.}
\label{tab:com_embedding_metric}
\begin{tabular}{|l|c|c|c|c|c|c|}
\toprule
\multirow{2}{*}{Methods} & \multicolumn{2}{c|}{Greedy@k} & \multicolumn{2}{c|}{Average@k} & \multicolumn{2}{c|}{Extrema@k} \\
                         & @10      & @100               & @10   & @100                   & @10    & @100     \\
\midrule                    

RNN          & 0.079      & 0.081& 0.365 & 0.373 & 0.147  & 0.147  \\
SA-BiRNNs    & 0.080      & 0.081& 0.385 & 0.390 & 0.152  & 0.152  \\
GRSM         & 0.102      & 0.104& 0.423 & 0.421 & 0.153  & 0.152  \\
Bert         & 0.106      & 0.101       & 0.414     & 0.399 & 0.207  & 0.196  \\
MSSNN        & \textbf{0.116}    & \textbf{0.109}& \textbf{0.439} & \textbf{0.425} & \textbf{0.218}  & \textbf{0.207}  \\       
\bottomrule
\end{tabular}
\end{table*}

\subsubsection{Evaluation} is conducted on large-scale responses selection. We use response set $R$ in the testing data (0.1 million for Douban and 18920 for Ubuntu) as index, and use query $\mathbf{q} \in Q$ to retrieve. Recall of ground-truth response can demonstrate the model's performance, so R@k and R-precision are calculated as metrics,
\begin{align}
\textnormal{R@k} &= \frac{1}{|Q|}\sum_{\mathbf{q} \in Q}{\frac{|\tilde{R}_{\mathbf{q}}\textnormal{@k}|}{|R_{\mathbf{q}}|}} \\
\textnormal{R-precision} &= \frac{1}{|Q|}\sum_{\mathbf{q} \in Q}{\frac{1}{\textnormal{position of }\mathbf{r_q}}}
\end{align}
where $|\tilde{R}_\mathbf{q}\textnormal{@k}|$ stands for count of recalled responses to query $\mathbf{q}$ in top k, $\mathbf{r_q}$ denotes true response to $\mathbf{q}$, $|R_\mathbf{q}|$ is count of ground truth responses to $\mathbf{q}$, and $|Q|$ means total queries count. In addition, because the selected responses are used to support the re-ranking, we also evaluate the quality of top-k responses using embedding-based metrics (EmM): greedy matching (Greedy), embedding average (Average), and vector extrema (Extrema). And we use EmM@k to represent the average scores of top-k selected responses,
\begin{align}
\textnormal{EmM@k} = \frac{1}{|Q|} \sum_{\mathbf{q}\in Q}{\frac{1}{\textnormal{k}}\sum_{\tilde{\mathbf{r}} \in \tilde{R}_\mathbf{q}\textnormal{@k}}{\textnormal{EmM}(\mathbf{r_q}, \tilde{\mathbf{r}})}}
\end{align}
where EmM is the alternate of the three metrics. 

\subsection{Results and analysis}\label{sec:results}

To test the ability of retrieving true response from large-scale corpus, experiments are first conducted on Douban and Ubuntu dataset using metrics of R-precision and R@k. Notice that since testing set of Ubuntu is smaller than Douban, we test R@100/R@1000 for Douban and R@10/R@100 for Ubuntu. The results are reported in Table \ref{tab:comarison}. Firstly, we find Lucene's results are worse than Bert or MSSNN because we found that there is no overlap in more than 67\% of QR pairs in Douban, and 65\% in Ubuntu dataset. Afterwards, It shows except for Bert's result on R-precision, our proposed method overcomes all other baseline methods. Bert also has a good results because of its pre-training by large corpus and large computation, but its architecture is still troubled with little interaction as well as selecting one truth response from a certain number of negative candidates, in which case it performs a little worse than MSSNN. And the left methods have bad results and fail in the QR semantic search task. Besides, we also evaluate the similarity between the true response and the top-k retrieval responses. Notice that in this experiment, we only use Douban dataset which is much larger than Ubuntu thereby calling for the more powerful semantic representations. The results are listed in Table \ref{tab:com_embedding_metric}. It illustrates that our method is the best from the corresponding metrics. 

In our method, Seq2Seq makes the encoder robust and gets strong ability in semantic representation of queries. In addition, Word Predictor is trained to predict all the words in corresponding responses, so that the weight $\mathbf{W}$ in Eq. \ref{eq:wordpred_w} bridges the gap of the semantic distance between the query and the response representations. Finally, benefiting from multi-task learning framework that combines them together with Matching Model, our model is proved to be the best within those baseline methods in QR semantic search regarding overall metrics. Besides, our model performs much better than Lucene index, and thereby it can improve the performance of retrieval-based chatbots.

\subsubsection{Response scale analysis}
\begin{figure}[!h]
  \centering
  \includegraphics[width=\linewidth]{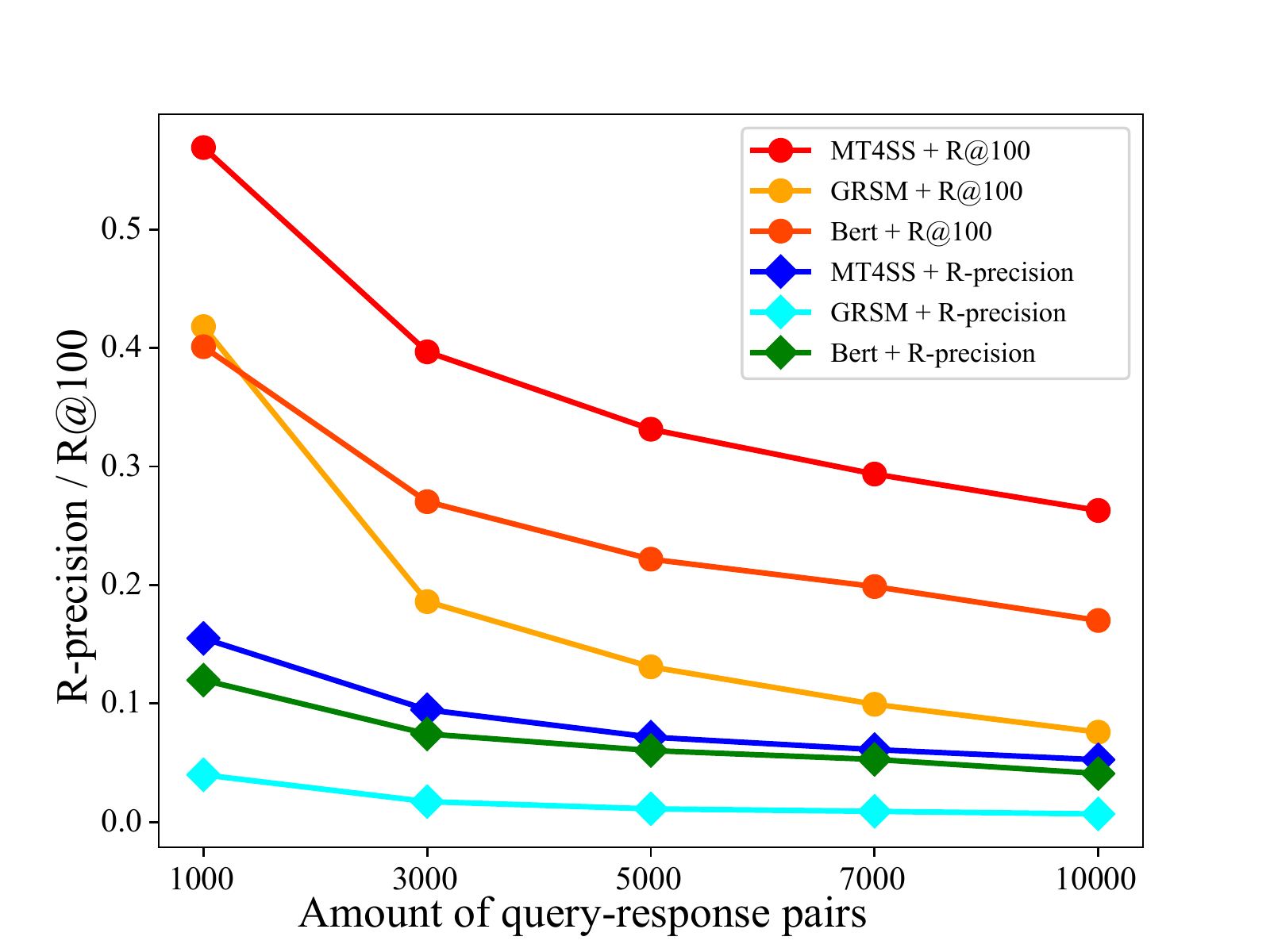}
  \caption{Response scale analysis within MSSNN, GRSM and Bert}
  \label{fig:data_amount}
\end{figure}

We randomly sample, from testing data of Douban, $(1, 3, 5, 7, 10)$ thousands query-response pairs to construct 5 testing sets, and we only compare MSSNN with GRSM and Bert because the other traditional methods fails in the semantic search task according to Table \ref{tab:comarison} and \ref{tab:com_embedding_metric}. With each testing set, we measure R-precision and R@100 to evaluate the influence of response scale on different methods. The results are reported in Fig. \ref{fig:data_amount} that X-axis denotes the amount of query-response pairs and the Y-axis means metric values. Notice that the above three lines are for R@100 and the others are for R-precision.

First it illustrates that the two metrics both decline with the growth of response scale, and it gives an expression that the larger the candidate response set is, the harder the method selects a true response from it. Moreover, our method MSSNN outperforms the other two methods with any response scale on both two metrics.

\subsubsection{Word Predictor's performance analysis}\label{sec:wppa}


Further experiment is conducted to analyze the performance of Word Predictor to highlight its function to our method. We sample 10 thousands queries from testing data in Douban dataset, and feed query $\mathbf{q}$ into MSSNN and yield the top-k predicted words by Word Predictor, $\tilde{V}_\mathbf{q}^{\textnormal{k}}$. Afterwards, we measure the coverage of predicted words in the ground truth response $\mathbf{r}$, $\textnormal{COV}_{\textnormal{@k}}$. In addition, we also measure its coverage in the generated response $\mathbf{r}_{\textnormal{gen}}$ by the Seq2Seq framework in MSSNN, $\textnormal{COV}_{\textnormal{@k}}^{\textnormal{gen}}$. The coverage is formalized as,
\begin{equation}\label{eq:coverage}
  \textnormal{COV}_{\textnormal{@k}} = \frac{1}{|Q|}\sum_{\mathbf{q}\in Q}{\frac{|\tilde{V}_\mathbf{q}^{\textnormal{k}} \cap V_{R_\mathbf{q}}|}{|V_{R_\mathbf{q}}|}}  
\end{equation}
where $V_{R_\mathbf{q}}$ is the word set of responses related to $\mathbf{q}$. Results are listed in Table \ref{tab:coverage}. It shows that the top-1000 predicted words from 30,000-words vocabulary can cover most words in generative or ground truth response. This result proves its capacity to predict the words in related responses solely based on a query. Section Word Predictor for Response Embedding tells us that Word Predictor utilizes simply a weight $\mathbf{W}$ in Eq. \ref{eq:wordpred_w} to map the query's semantic space $\mathbf{v_q}$ to the response's vocabulary space. To this point, it demonstrates the appropriateness of using transposed $\mathbf{W}$ as word embedding to map response words to query's semantic space, and it improves the performance of query-response semantic search.

\begin{table}[!h]
\centering
\caption{Word Predictor: Coverage of top-k predicted words in responses, vocabulary size equals to 30,000.}
\label{tab:coverage}
\begin{tabular}{|l|c|c|c|}
\hline
top k      &  1,000     &  5,000    &  10,000 \\
\hline
$\textnormal{COV}_{\textnormal{@k}}$        &  0.808  &  0.942   &  0.971  \\
\hline
$\textnormal{COV}_{\textnormal{@k}}^{\textnormal{gen}}$  &  0.945  &  0.993   &  0.998  \\
\hline
\end{tabular}
\end{table}

\subsubsection{First-step prediction in Seq2Seq}\label{sec:fsp}
Auxiliary experiment on first-step decoding in Seq2Seq is conducted to verify the effectiveness of involving the Seq2Seq task. We use the same Seq2Seq architecture with the Seq2Seq in MSSNN, and use the same Douban conversational dataset. After 10 iterations, the training/validation PPL is descending to 66.47/70.86. Afterwards, we test the first-step prediction on 10000 queries and calculate the coverage of predicting top-k words from corresponding responses, ground truth response $\textnormal{COV}_{\textnormal{@k}}$ and generated response $\mathbf{r}_{\textnormal{gen}}$. Results listed in Table \ref{tab:coverage_s2s} have shown that the top-1000 predicted words from the vocabulary also can cover most words, which shares the similar results with previous research in \cite{DBLP:conf/aaai/WuWYXL18}. This result proves that the encoder in Seq2Seq achieves a good mapping from queries to the responses representations.

\begin{table}[!h]
\centering
\caption{First-step prediction in Seq2Seq: Coverage of top-k predicted words in responses}
\label{tab:coverage_s2s}
\begin{tabular}{|l|c|c|c|}
\hline
top k      &  1,000    &   5,000   &  10,000 \\
\hline
$\textnormal{COV}_{\textnormal{@k}}$        &  0.722   &  0.925  &  0.963   \\
\hline
$\textnormal{COV}_{\textnormal{@k}}^{\textnormal{gen}}$  &  0.840  & 0.977  &  0.993  \\
\hline
\end{tabular}
\end{table}

\subsubsection{Latency analysis}

In addition to fetching relevant responses for a candidate-retrieval module, another key point is the latency. Thus, we analyze the latency of the semantic search by comparing with Lucene. For semantic search, we apply the MSSNN/Bert to produce vectors of 100,000 sentences, and use Annoy 
 to build a forest of 400 trees and to approximate dot-producting search, MSSNN-512-ann/Bert-768-ann. As a comparison, we build Luence index on the same 100,000 sentences. The average latency of retrieving top-K candidates per question is listed in Table \ref{tab:latency}. Notes that they are executed in same environment. 
 
 \begin{table}[ht]
    \centering
    \caption{Latency analysis}
    \begin{tabular}{|c|c|c|c|}
    \hline
         Top-K & Lucene & MSSNN-512-ann & Bert-768-ann \\
         \hline
         10  & 2.66 &  1.35 & 1.78  \\
         \hline
         30  & 3.09 & 2.40 & 2.672  \\
         \hline
         \textbf{50}  & 3.38 & \textbf{3.29} & \textbf{3.38}  \\
         \hline
         70  & 3.54 & 4.12 & 4.17  \\
         \hline
         90  & 3.67 & 4.91 & 5.02  \\
         \hline
         110 & 3.78 & 5.63 & 5.76  \\
         \hline
         130 & 3.91 & 6.32 & 6.39 \\
         \hline
         150 & 4.02 & 6.95 & 7.01 \\
         \hline
         170 & 4.09 & 7.63 & 7.64 \\
         \hline
         190 & 4.15 & 8.21 & 8.23 \\
         \hline
    \end{tabular}
    
    \label{tab:latency}
\end{table}

    

Table \ref{tab:latency} has shown that the latency of semantic search is comparable to that of Lucene, or even semantic searching top-50 candidates from 100,000 using ANN tools is faster than Lucene index. The average latency of ANN augments faster than Lucene related to top-K searching. Normally, we would like to retrieve less than top-100 candidates in the first stage in chatbots, and these candidates are used for the second stage, which will use a more powerful re-ranking model. Therefore, the comparable latency of Lucene and ANN ensures the application of semantic search in chatbots.

\subsubsection{Case study}

\begin{table*}[!ht]
\scriptsize
\centering
\caption{Case study on top 5 responses selected from 100,000 response set.}
\label{tab:case_study}
\begin{tabularx}{\textwidth}{p{0.06cm}|p{5.3cm}|p{5.6cm}|p{3.5cm}}
\hline
    &\multicolumn{1}{c|}{MSSNN}  &  \multicolumn{1}{c|}{Bert}  & \multicolumn{1}{c}{GRSM}  \\
\hline
$\mathbf{q}$  & \multicolumn{3}{c}{\Zh{很古老的话题，人活着的意义是什么} (A very old topic, what is the meaning of being alive)} \\
\hline
 $\tilde{\mathbf{r}}_1$     & \Zh{寻找平静和信仰} (Looking for peace and faith)  & \Zh{关系太多也会束缚人生} (Life will be bounded by too many relationships)  &  \Zh{为什么分手} (Why broke up) \\
\hline
 $\tilde{\mathbf{r}}_2$     & \Zh{因果报应轮回，每个人都是一样的} (The reincarnation of Karma is the same for everyone)  &  \Zh{问一个羞涩的问题，狗狗什么性别} (Ask an embarrassed question - what is the gender of the dog) &  \Zh{贱人就是矫情} (Bitch is hypocritical) \\
\hline
 $\tilde{\mathbf{r}}_3$     & \Zh{理解每个人心里都住着一个哲学家} (You should understand that there lives a philosopher in every one's heart)  & \Zh{这就是灵魂，人死了就是这样} (It's a soul. That's what happends when people die)  &  \Zh{让他滚} (Let him get out) \\
\hline
 $\tilde{\mathbf{r}}_4$     & \Zh{自己就应该活的更精彩} (One should live a better life)  &  \Zh{如果人生没有梦想，那和咸鱼有什么分别} (A man without dream makes no differences with a salted fish) &  \Zh{求骂醒} (Please scold and wake me up) \\
\hline
 $\tilde{\mathbf{r}}_5$     & \Zh{有时候感觉自己与这个世界格格不入} (Sometimes, I feel like I am out of tune with the world)  & \Zh{快乐是什么味道} (What is the taste of happiness) &  \Zh{女人心海底针} (A woman's heart is as deep as the ocean) \\
\hline
\end{tabularx}

\begin{tabularx}{\textwidth}{c}
 \\
\end{tabularx}

\begin{tabularx}{\textwidth}{p{0.06cm}|p{5.3cm}|p{5.6cm}|p{3.5cm}}
\hline
$\mathbf{q}$  & \multicolumn{3}{c}{\Zh{在干嘛} (What are you doing)} \\
\hline
$\tilde{\mathbf{r}}_1$  & \Zh{回家还刷豆瓣，今天的确很爽} (Going home and surfing the Douban forum, today is really cool)  & \Zh{来干嘛} (What are you coming for)  &  \Zh{晚安，好梦} (Good night, with sweet dreams) \\
\hline
$\tilde{\mathbf{r}}_2$  & \Zh{人家在上班啦} (I am at work)  &  \Zh{干嘛呀} (What's up) &  \Zh{我没睡} (I have not slept yet) \\
\hline
$\tilde{\mathbf{r}}_3$  & \Zh{你下午不忙} (You are not busy in the afternoon)  &  \Zh{干嘛啦} (What are you doing) &  \Zh{晚睡强迫症} (Obsessive thoughts of forcing to sleep late) \\
\hline
$\tilde{\mathbf{r}}_4$  & \Zh{太无聊了,等下班放假了} (It's so boring, i am waiting for the holiday after get off work)  & \Zh{在做啊} (I am doing now)  &  \Zh{你没睡} (You did not sleep) \\
\hline
$\tilde{\mathbf{r}}_5$  & \Zh{刚回学校就在寝室里上网} (Surf the Internet at the dormitory when going back to school)  & \Zh{干什么呀} (What are you doing) &  \Zh{出去玩吧} (Hang out and play) \\
\hline
\end{tabularx}

\begin{tabularx}{\textwidth}{c}
 \\
\end{tabularx}

\begin{tabularx}{\textwidth}{p{0.06cm}|p{5.3cm}|p{5.6cm}|p{3.5cm}}
\hline
$\mathbf{q}$  & \multicolumn{3}{c}{\Zh{周末有什么好玩的么} (Is there anything fun on the weekend)} \\
\hline
$\tilde{\mathbf{r}}_1$  & \Zh{正在双井无聊中} (Being bored in Shuangjing)  &  \Zh{有什么玩的} (What's fun) &  \Zh{什么时候出发} (When to set off) \\
\hline
$\tilde{\mathbf{r}}_2$  & \Zh{看电影吃饭，逛城墙都可以} (Watching movies, eating food, visiting the city walls)  &  \Zh{那里好玩儿吗} (Is it fun there)  &  \Zh{从哪里出发} (Where can we set off) \\
\hline
$\tilde{\mathbf{r}}_3$  & \Zh{去啊，陪玩} (Go and i will play with you)  & \Zh{有空骑车啊，周末怎么安排} (Take a bike ride if got time, and what's the plan for the weekend) &  \Zh{我在卢旺达} (I am in Rwanda) \\
\hline
$\tilde{\mathbf{r}}_4$  & \Zh{知道就好，你可以选择夜探香山} (It's alright if you know, you can choose to explore Xiangshan at night)  &  \Zh{谁能告诉我深圳有什么好玩的} (Can anyone tell me what is fun in Shenzhen) &  \Zh{西安欢迎你} (Welcome to Xi'an) \\
\hline
$\tilde{\mathbf{r}}_5$  & \Zh{上哪high去啊，在家high不了} (Where can I play, i can’t play at home)  & \Zh{10月份厦门什么天气啊，适合去玩么} (What's the weather in Xiamen in October, is it suitable for traveling)  &  \Zh{几号到} (What's the date of your arriving) \\
\hline
\end{tabularx}
\end{table*}

We conduct case study by using different methods to select top 5 responses from 100,000 response on Douban dataset. Notice that we only list the results of GRSM, Bert, and MSSNN, because the other methods retrieve totally irrelevant responses. Results are listed in Table \ref{tab:case_study}, where $\tilde{\mathbf{r}}_i$ is top $i$ retrieval response by the respective models. It illustrates that most selected responses by GRSM have bad relevance with the queries, and Bert's responses are more relevant to the queries but sentences which are similar to the query are more likely to be selected. In contrast, MSSNN can provide appropriate responses considering both the diversity and the relevance, which is valuable for the following re-ranking system and finally promotes the quality of the returned responses to users in chatbots.

\section{Related Works}
\subsubsection{Retrieval based Chatbots} can be divided into two parts, retrieve module and re-ranking module. Re-ranking module~\cite{DBLP:conf/ijcai/WanLXGPC16,DBLP:conf/emnlp/ZhouDWZYTLY16,DBLP:conf/acl/WuLCZDYZL18,DBLP:journals/corr/abs-2004-03588}, which selects the most appropriate response from some candidates attracted more attention than retrieve module, which usually uses Lucene engine to retrieve candidates from large corpus. In addition to Lucene, semantic search is an another solution, which firstly encodes queries and responses into vectors with same semantic space, and then employs nearest neighbor search (NNS) to do vector-search-vector in the same semantic space. Many approximative algorithms~\cite{DBLP:conf/iccv/ShenLZYS15,DBLP:conf/uai/Shrivastava015,DBLP:journals/corr/AuvolatV15} and existing tools can be referred, so our work focus on how to encode queries and responses into vectors with same semantic space.


\subsubsection{Matching models} for learning QR semantic relevance can be roughly divided into two categories: matching function learning and representation learning. 1) Methods of matching function learning first use handcrafted features or shallow models to represent respectively queries and responses, and then apply some deep models to discover the matching patterns, for example, ARC-\uppercase\expandafter{\romannumeral2}~\cite{DBLP:conf/nips/HuLLC14}, MatchPyramid~\cite{DBLP:conf/aaai/PangLGXWC16}, Match-SRNN~\cite{DBLP:conf/ijcai/WanLXGPC16}, RCNN~\cite{DBLP:conf/www/PengCXLZL20}, SA-BERT~\cite{DBLP:journals/corr/abs-2004-03588}. Their characteristic is that the query and the response interacts at the beginning which causes huge time cost for online calculation. 2) Methods of representation learning first use deep models to model and calculate representations for query and response, and finally use simple matching function like dot-product and cosine distance to score their relevance~\cite{DBLP:conf/ijcai/QiuH15,DBLP:conf/acl/WangN15,DBLP:journals/taslp/PalangiDSGHCSW16}. Representation learned by these methods work well for re-ranking task but fail in the semantic search task who has large-scale responses. Some recent works based on Bert also try to apply semantic search in some areas, like similar text matching \cite{DBLP:conf/emnlp/ReimersG19} or open-domain question answering \cite{DBLP:conf/acl/LeeCT19}, which are also applied as our baselines. In our work, we also use matching model to form the dot-product scoring. But the difference is that semantic representations are learned not only by matching model but also by two extra tasks who have powerful ability to map queries and responses into same semantic distance. 

\subsubsection{Multi-task Learning} is widely used to improve generalization on the target task by leveraging information contained tasks \cite{DBLP:books/sp/98/Caruana98}, including nature language processing \cite{DBLP:journals/tacl/PengPQTY17}, speech recognition \cite{DBLP:journals/corr/ThandaV17}, web search \cite{DBLP:conf/iclr/AhmadCW18}, and computer vision \cite{DBLP:conf/iccv/Girshick15}. We proposed to use multi-task learning to emphasize the interactions between queries and responses, and to construct query-response semantic search in chatbots.

\section{Conclusions}

This paper targets at finding out a solution of using query-response semantic search for candidate responses retrieval in chatbots. In this regard, we try to construct two mapping functions $\phi$ and $\psi$ with keeping in mind that the interaction between the query and response is important. Then inspired by multi-task learning framework, we proposed a novel approach MSSNN and final experimental studies have verified its effectiveness.

\bibliographystyle{aaai21}
\bibliography{aaai21}

\begin{thebibliography}{40}
\providecommand{\natexlab}[1]{#1}
\providecommand{\url}[1]{\texttt{#1}}
\providecommand{\urlprefix}{URL }
\expandafter\ifx\csname urlstyle\endcsname\relax
  \providecommand{\doi}[1]{doi:\discretionary{}{}{}#1}\else
  \providecommand{\doi}{doi:\discretionary{}{}{}\begingroup
  \urlstyle{rm}\Url}\fi

\bibitem[{Ahmad, Chang, and Wang(2018)}]{DBLP:conf/iclr/AhmadCW18}
Ahmad, W.~U.; Chang, K.; and Wang, H. 2018.
\newblock Multi-Task Learning for Document Ranking and Query Suggestion.
\newblock In \emph{{ICLR}}. OpenReview.net.

\bibitem[{Auvolat and Vincent(2015)}]{DBLP:journals/corr/AuvolatV15}
Auvolat, A.; and Vincent, P. 2015.
\newblock Clustering is Efficient for Approximate Maximum Inner Product Search.
\newblock \emph{CoRR} abs/1507.05910.

\bibitem[{Bahdanau, Cho, and Bengio(2014)}]{DBLP:journals/corr/BahdanauCB14}
Bahdanau, D.; Cho, K.; and Bengio, Y. 2014.
\newblock Neural Machine Translation by Jointly Learning to Align and
  Translate.
\newblock \emph{CoRR} abs/1409.0473.

\bibitem[{Caruana(1998)}]{DBLP:books/sp/98/Caruana98}
Caruana, R. 1998.
\newblock Multitask Learning.
\newblock In Thrun, S.; and Pratt, L.~Y., eds., \emph{Learning to Learn},
  95--133. Springer.

\bibitem[{Cho et~al.(2014)Cho, van Merrienboer, Bahdanau, and
  Bengio}]{DBLP:conf/ssst/ChoMBB14}
Cho, K.; van Merrienboer, B.; Bahdanau, D.; and Bengio, Y. 2014.
\newblock On the Properties of Neural Machine Translation: Encoder-Decoder
  Approaches.
\newblock In \emph{Proc. of {SSST@EMNLP}}.

\bibitem[{Collobert and Weston(2008)}]{DBLP:conf/icml/CollobertW08}
Collobert, R.; and Weston, J. 2008.
\newblock A unified architecture for natural language processing: deep neural
  networks with multitask learning.
\newblock In \emph{Proc. of {ICML}}, 160--167.

\bibitem[{Devlin et~al.(2018)Devlin, Chang, Lee, and
  Toutanova}]{DBLP:journals/corr/abs-1810-04805}
Devlin, J.; Chang, M.; Lee, K.; and Toutanova, K. 2018.
\newblock {BERT:} Pre-training of Deep Bidirectional Transformers for Language
  Understanding.
\newblock \emph{CoRR} abs/1810.04805.

\bibitem[{Girshick(2015)}]{DBLP:conf/iccv/Girshick15}
Girshick, R.~B. 2015.
\newblock Fast {R-CNN}.
\newblock In \emph{{ICCV}}, 1440--1448. {IEEE} Computer Society.

\bibitem[{Gu et~al.(2020)Gu, Li, Liu, Zhu, Ling, Su, and
  Wei}]{DBLP:journals/corr/abs-2004-03588}
Gu, J.; Li, T.; Liu, Q.; Zhu, X.; Ling, Z.; Su, Z.; and Wei, S. 2020.
\newblock Speaker-Aware {BERT} for Multi-Turn Response Selection in
  Retrieval-Based Chatbots.
\newblock \emph{CoRR} abs/2004.03588.

\bibitem[{Henderson et~al.(2017)Henderson, Al{-}Rfou, Strope, Sung,
  Luk{\'{a}}cs, Guo, Kumar, Miklos, and
  Kurzweil}]{DBLP:journals/corr/HendersonASSLGK17}
Henderson, M.; Al{-}Rfou, R.; Strope, B.; Sung, Y.; Luk{\'{a}}cs, L.; Guo, R.;
  Kumar, S.; Miklos, B.; and Kurzweil, R. 2017.
\newblock Efficient Natural Language Response Suggestion for Smart Reply.
\newblock \emph{CoRR} abs/1705.00652.

\bibitem[{Hu et~al.(2014)Hu, Lu, Li, and Chen}]{DBLP:conf/nips/HuLLC14}
Hu, B.; Lu, Z.; Li, H.; and Chen, Q. 2014.
\newblock Convolutional Neural Network Architectures for Matching Natural
  Language Sentences.
\newblock In \emph{Proc. of {NIPS}}, 2042--2050.

\bibitem[{Ji, Lu, and Li(2014)}]{DBLP:journals/corr/JiLL14}
Ji, Z.; Lu, Z.; and Li, H. 2014.
\newblock An Information Retrieval Approach to Short Text Conversation.
\newblock \emph{CoRR} abs/1408.6988.

\bibitem[{Kingma and Ba(2014)}]{DBLP:journals/corr/KingmaB14}
Kingma, D.~P.; and Ba, J. 2014.
\newblock Adam: {A} Method for Stochastic Optimization.
\newblock \emph{CoRR} abs/1412.6980.

\bibitem[{Lee, Chang, and Toutanova(2019)}]{DBLP:conf/acl/LeeCT19}
Lee, K.; Chang, M.; and Toutanova, K. 2019.
\newblock Latent Retrieval for Weakly Supervised Open Domain Question
  Answering.
\newblock In Korhonen, A.; Traum, D.~R.; and M{\`{a}}rquez, L., eds.,
  \emph{Proc. of {ACL}}, 6086--6096.

\bibitem[{Li et~al.(2017)Li, Qiu, Chen, Wang, Gao, Huang, Ren, Zhao, Zhao,
  Wang, Jin, and Chu}]{DBLP:conf/cikm/LiQCWGHRZZWJC17}
Li, F.; Qiu, M.; Chen, H.; Wang, X.; Gao, X.; Huang, J.; Ren, J.; Zhao, Z.;
  Zhao, W.; Wang, L.; Jin, G.; and Chu, W. 2017.
\newblock \emph{AliMe Assist }: An Intelligent Assistant for Creating an
  Innovative E-commerce Experience.
\newblock In \emph{Proc. of {CIKM}}, 2495--2498.

\bibitem[{Lowe et~al.(2015)Lowe, Pow, Serban, and
  Pineau}]{DBLP:conf/sigdial/LowePSP15}
Lowe, R.; Pow, N.; Serban, I.; and Pineau, J. 2015.
\newblock The Ubuntu Dialogue Corpus: {A} Large Dataset for Research in
  Unstructured Multi-Turn Dialogue Systems.
\newblock In \emph{Proc. of the {SIGDIAL}}, 285--294.

\bibitem[{Mikolov, Le, and Sutskever(2013)}]{DBLP:journals/corr/MikolovLS13}
Mikolov, T.; Le, Q.~V.; and Sutskever, I. 2013.
\newblock Exploiting Similarities among Languages for Machine Translation.
\newblock \emph{CoRR} abs/1309.4168.

\bibitem[{Mikolov et~al.(2013)Mikolov, Sutskever, Chen, Corrado, and
  Dean}]{DBLP:conf/nips/MikolovSCCD13}
Mikolov, T.; Sutskever, I.; Chen, K.; Corrado, G.~S.; and Dean, J. 2013.
\newblock Distributed Representations of Words and Phrases and their
  Compositionality.
\newblock In \emph{Proc. of {NIPS}}, 3111--3119.

\bibitem[{Palangi et~al.(2016)Palangi, Deng, Shen, Gao, He, Chen, Song, and
  Ward}]{DBLP:journals/taslp/PalangiDSGHCSW16}
Palangi, H.; Deng, L.; Shen, Y.; Gao, J.; He, X.; Chen, J.; Song, X.; and Ward,
  R.~K. 2016.
\newblock Deep Sentence Embedding Using Long Short-Term Memory Networks:
  Analysis and Application to Information Retrieval.
\newblock \emph{{IEEE/ACM} Trans. Audio, Speech {\&} Language Processing}
  24(4): 694--707.

\bibitem[{Pang et~al.(2016)Pang, Lan, Guo, Xu, Wan, and
  Cheng}]{DBLP:conf/aaai/PangLGXWC16}
Pang, L.; Lan, Y.; Guo, J.; Xu, J.; Wan, S.; and Cheng, X. 2016.
\newblock Text Matching as Image Recognition.
\newblock In \emph{Proc. of {AAAI}}, 2793--2799.

\bibitem[{Peng et~al.(2017)Peng, Poon, Quirk, Toutanova, and
  Yih}]{DBLP:journals/tacl/PengPQTY17}
Peng, N.; Poon, H.; Quirk, C.; Toutanova, K.; and Yih, W. 2017.
\newblock Cross-Sentence N-ary Relation Extraction with Graph LSTMs.
\newblock \emph{Trans. Assoc. Comput. Linguistics} 5: 101--115.

\bibitem[{Peng et~al.(2020)Peng, Cui, Xie, Li, Zhang, and
  Li}]{DBLP:conf/www/PengCXLZL20}
Peng, S.; Cui, H.; Xie, N.; Li, S.; Zhang, J.; and Li, X. 2020.
\newblock Enhanced-RCNN: An Efficient Method for Learning Sentence Similarity.
\newblock In Huang, Y.; King, I.; Liu, T.; and van Steen, M., eds.,
  \emph{{WWW}}, 2500--2506.

\bibitem[{Qiu and Huang(2015)}]{DBLP:conf/ijcai/QiuH15}
Qiu, X.; and Huang, X. 2015.
\newblock Convolutional Neural Tensor Network Architecture for Community-Based
  Question Answering.
\newblock In \emph{Proc. of {IJCAI}}, 1305--1311.

\bibitem[{Reimers and Gurevych(2019)}]{DBLP:conf/emnlp/ReimersG19}
Reimers, N.; and Gurevych, I. 2019.
\newblock Sentence-BERT: Sentence Embeddings using Siamese BERT-Networks.
\newblock In Inui, K.; Jiang, J.; Ng, V.; and Wan, X., eds., \emph{Proc. of
  {EMNLP-IJCNLP}}, 3980--3990.

\bibitem[{R{\"{u}}ckl{\'{e}} and Gurevych(2017)}]{DBLP:conf/iwcs/RuckleG17}
R{\"{u}}ckl{\'{e}}, A.; and Gurevych, I. 2017.
\newblock Representation Learning for Answer Selection with LSTM-Based
  Importance Weighting.
\newblock In \emph{Proc. of {IWCS}}.

\bibitem[{Serban et~al.(2016)Serban, Sordoni, Bengio, Courville, and
  Pineau}]{DBLP:conf/aaai/SerbanSBCP16}
Serban, I.~V.; Sordoni, A.; Bengio, Y.; Courville, A.~C.; and Pineau, J. 2016.
\newblock Building End-To-End Dialogue Systems Using Generative Hierarchical
  Neural Network Models.
\newblock In \emph{Proc. of {AAAI}}, 3776--3784.

\bibitem[{Shang, Lu, and Li(2015)}]{DBLP:conf/acl/ShangLL15}
Shang, L.; Lu, Z.; and Li, H. 2015.
\newblock Neural Responding Machine for Short-Text Conversation.
\newblock In \emph{Proc. of {ACL}}, 1577--1586.

\bibitem[{Shen et~al.(2015)Shen, Liu, Zhang, Yang, and
  Shen}]{DBLP:conf/iccv/ShenLZYS15}
Shen, F.; Liu, W.; Zhang, S.; Yang, Y.; and Shen, H.~T. 2015.
\newblock Learning Binary Codes for Maximum Inner Product Search.
\newblock In \emph{Proc. of {ICCV}}, 4148--4156.

\bibitem[{Shrivastava and Li(2015)}]{DBLP:conf/uai/Shrivastava015}
Shrivastava, A.; and Li, P. 2015.
\newblock Asymmetric Locality Sensitive Hashing {(ALSH)} for Maximum Inner
  Product Search {(MIPS)}.
\newblock In \emph{Proc. of {UAI}}, 812--821.

\bibitem[{Tan, Xiang, and Zhou(2015)}]{DBLP:journals/corr/TanXZ15}
Tan, M.; Xiang, B.; and Zhou, B. 2015.
\newblock LSTM-based Deep Learning Models for non-factoid answer selection.
\newblock \emph{CoRR} abs/1511.04108.

\bibitem[{Thanda and Venkatesan(2017)}]{DBLP:journals/corr/ThandaV17}
Thanda, A.; and Venkatesan, S.~M. 2017.
\newblock Multi-task Learning Of Deep Neural Networks For Audio Visual
  Automatic Speech Recognition.
\newblock \emph{CoRR} abs/1701.02477.

\bibitem[{Wan et~al.(2016)Wan, Lan, Xu, Guo, Pang, and
  Cheng}]{DBLP:conf/ijcai/WanLXGPC16}
Wan, S.; Lan, Y.; Xu, J.; Guo, J.; Pang, L.; and Cheng, X. 2016.
\newblock Match-SRNN: Modeling the Recursive Matching Structure with Spatial
  {RNN}.
\newblock In \emph{Proc. of {IJCAI}}, 2922--2928.

\bibitem[{Wang and Nyberg(2015)}]{DBLP:conf/acl/WangN15}
Wang, D.; and Nyberg, E. 2015.
\newblock A Long Short-Term Memory Model for Answer Sentence Selection in
  Question Answering.
\newblock In \emph{Proc. of {ACL}}, 707--712.

\bibitem[{Wu et~al.(2017)Wu, Wu, Xing, Zhou, and Li}]{DBLP:conf/acl/WuWXZL17}
Wu, Y.; Wu, W.; Xing, C.; Zhou, M.; and Li, Z. 2017.
\newblock Sequential Matching Network: {A} New Architecture for Multi-turn
  Response Selection in Retrieval-Based Chatbots.
\newblock In \emph{Proc. of {ACL}}, 496--505.

\bibitem[{Wu et~al.(2018)Wu, Wu, Yang, Xu, and Li}]{DBLP:conf/aaai/WuWYXL18}
Wu, Y.; Wu, W.; Yang, D.; Xu, C.; and Li, Z. 2018.
\newblock Neural Response Generation With Dynamic Vocabularies.
\newblock In McIlraith, S.~A.; and Weinberger, K.~Q., eds., \emph{Proc. of
  {AAAI}}, 5594--5601.

\bibitem[{Xing et~al.(2018)Xing, Wu, Wu, Huang, and
  Zhou}]{DBLP:conf/aaai/XingWWHZ18}
Xing, C.; Wu, Y.; Wu, W.; Huang, Y.; and Zhou, M. 2018.
\newblock Hierarchical Recurrent Attention Network for Response Generation.
\newblock In \emph{Proc. of {AAAI}}, 5610--5617.

\bibitem[{Yan, Song, and Wu(2016)}]{DBLP:conf/sigir/YanSW16}
Yan, R.; Song, Y.; and Wu, H. 2016.
\newblock Learning to Respond with Deep Neural Networks for Retrieval-Based
  Human-Computer Conversation System.
\newblock In \emph{Proc. of {SIGIR}}, 55--64.

\bibitem[{Zhou et~al.(2018{\natexlab{a}})Zhou, Gao, Li, and
  Shum}]{DBLP:journals/corr/abs-1812-08989}
Zhou, L.; Gao, J.; Li, D.; and Shum, H. 2018{\natexlab{a}}.
\newblock The Design and Implementation of XiaoIce, an Empathetic Social
  Chatbot.
\newblock \emph{CoRR} abs/1812.08989.

\bibitem[{Zhou et~al.(2016)Zhou, Dong, Wu, Zhao, Yu, Tian, Liu, and
  Yan}]{DBLP:conf/emnlp/ZhouDWZYTLY16}
Zhou, X.; Dong, D.; Wu, H.; Zhao, S.; Yu, D.; Tian, H.; Liu, X.; and Yan, R.
  2016.
\newblock Multi-view Response Selection for Human-Computer Conversation.
\newblock In \emph{Proc. of {EMNLP}}, 372--381.

\bibitem[{Zhou et~al.(2018{\natexlab{b}})Zhou, Li, Dong, Liu, Chen, Zhao, Yu,
  and Wu}]{DBLP:conf/acl/WuLCZDYZL18}
Zhou, X.; Li, L.; Dong, D.; Liu, Y.; Chen, Y.; Zhao, W.~X.; Yu, D.; and Wu, H.
  2018{\natexlab{b}}.
\newblock Multi-Turn Response Selection for Chatbots with Deep Attention
  Matching Network.
\newblock In Gurevych, I.; and Miyao, Y., eds., \emph{Proc. of {ACL}},
  1118--1127.

\end{thebibliography}

\end{document}